\documentclass[lettersize,journal]{IEEEtran}
\usepackage{amsmath,amsfonts}
\usepackage{algorithmic}
\usepackage{algorithm}
\usepackage{array}
\usepackage[caption=false,font=normalsize,labelfont=sf,textfont=sf]{subfig}
\usepackage{textcomp}
\usepackage{stfloats}
\usepackage{url}
\usepackage{verbatim}
\usepackage{graphicx}
\usepackage{cite}
\usepackage{bm}
\usepackage{multirow}
\UseRawInputEncoding
\newcommand{\upcite}[1]{$^{\mbox{\scriptsize \cite{#1}}}$}
\newcommand{\tabincell}[2]{\begin{tabular}{@{}#1@{}}#2\end{tabular}}
\begin{document}

\title{HRPose: Real-Time High-Resolution 6D Pose Estimation Network Using Knowledge Distillation}

\author{Qi Guan, Zihao Sheng, and Shibei Xue~\IEEEmembership{Senior Member IEEE}\\ 
\thanks{Q. Guan, Z. H. Sheng, and S. Xue are with Department of Automation, Shanghai
Jiao Tong University, Shanghai, 200240, P. R. China; Key Laboratory of System Control and Information Processing,  Ministry of Education of China, Shanghai 200240, P. R. China; Shanghai Engineering Research Center of Intelligent Control and Management, Shanghai 200240, P. R. China (e-mail: {shbxue}@sjtu.edu.cn).}
\thanks{This work was supported by the National Natural Science Foundation of China (61873162, 61973317) and Open Research Project of the State Key Laboratory of Industrial Control Technology, Zhejiang University, China (ICT2022B47).}}


\IEEEpubid{10.1049/cje.2021.00.211~\copyright~2022 Chinese Insitute of Electronics}

\maketitle

\begin{abstract}
Real-time 6D object pose estimation is essential for many real-world applications, such as robotic grasping and augmented reality. To achieve an accurate object pose estimation from RGB images in real-time, we propose an effective and lightweight model, namely High-Resolution 6D Pose Estimation Network (HRPose). We adopt the efficient and small HRNetV2-W18 as a feature extractor to reduce computational burdens while generating accurate 6D poses. With only 33\% of the model size and lower computational costs, our HRPose achieves comparable performance compared with state-of-the-art models. Moreover, by transferring knowledge from a large model to our proposed HRPose through output and feature-similarity distillations, the performance of  our HRPose is improved in effectiveness and efficiency. Numerical experiments on the widely-used benchmark LINEMOD demonstrate the superiority of our proposed HRPose against state-of-the-art methods.
\end{abstract}


\section{Introduction}
Object pose estimation aims to obtain the 6DoF (6 degrees of freedom) pose of an object in a camera coordinate, and its real-time application is crucial for autonomous driving\upcite{xu2018pointfusion,sheng2020real,sheng2022cooperation,sheng2022graph}, augmented reality\upcite{marchand2015pose}, robotic grasping\upcite{Yu2018PoseCNN}, and so forth. For instance, fast and accurate 6D pose estimation is essential in Amazon Picking Challenge\upcite{AmazonChallenge}, where a robot needs to pick objects from a warehouse shelf. Although methods\upcite{wang2019densefusion,he2020pvn3d} that rely on depth images for this task are more robust, estimating object poses from RGB images is more attractive for actual scenarios, in terms of hardware cost and availability. This problem is still challenging due to variations in appearance and cluttered environments.

Traditional methods\upcite{lowe1999object,hinterstoisser2011gradient} compute object poses by establishing maps between object images and their actual model through feature points or template matching. They rely on hand-crafted features, which are sensitive to image variations and background clutters. Nowadays, with the development of deep learning, deep Convolutional Neural Networks (CNNs) have achieved significant progresses in 6D object pose estimation.

To achieve an efficient pose estimation, existing methods\upcite{rad2017bb8,tekin2018real,oberweger2018making,Peng2020pvnet} first use CNNs to detect predefined 2D keypoints and then recover object poses via a Perspective-n-Point (PnP) algorithm\upcite{lepetit2009epnp}. Among these methods, Tekin \textit{et al.}\upcite{tekin2018real} employed the object detector YOLOv2\upcite{redmon2017yolo9000} to directly regress 2D locations of keypoints which can achieve almost the fastest speed in pose estimation. However, directly regressing the keypoints coordinates makes the CNN hard to converge, which results in degradation of accuracy. Tremblay \textit{et al.}\upcite{tremblay2018deep} proposed a multistage architecture to estimate pixel-wise heatmaps of 2D keypoints. Peng \textit{et al.}\upcite{Peng2020pvnet} proposed pixel-wise unit vectors as a representation of keypoints. However, such dense predictions lead to an increase in model size and computational complexity, which restricts them from actual applications.

\IEEEpubidadjcol
Recently, neural networks with high accuracies, small model sizes, and low computational costs have attracted much attention for their demands in resource-limited devices, such as embedded systems\upcite{du2021vision}. For such systems, knowledge distillation\upcite{hinton2015distilling} has been widely studied for its simplicity and effectiveness. Its main idea is to improve the performance of a small network by transferring the knowledge from a large teacher network. Felix \textit{et al.}\upcite{KD2020pose} proposed an improved knowledge distillation to get a faster version of a 2D keypoints detector based on YOLO6D\upcite{tekin2018real} for object pose estimation, where the output of a trained teacher network is simply transfer . However, the unreliable teacher network introduces noises in training and makes the student network fail to meet requirements on accuracies in real-world applications.

An ideal solution to 6D object pose estimation should satisfy some actual conditions such as textureless appearance, heavy clutter scenes, and environmental variations. Also, it should meet the speed requirement for real-time tasks (e.g. 30 frames per second\upcite{redmon2017yolo9000}). To this end, we propose a simple and efficient model, namely High-Resolution 6D Pose Estimation Network (HRPose) which predicts keypoints from the high-resolution feature representation in a bottom-up method. HRPose takes the small HRNetV2-W18\upcite{wang2020deep} as the backbone and is able to retain spatial positions as well as deep semantic information, which leads to more accurate pose estimations.

To further improve the estimation accuracy of HRPose without a performance degradation, we propose a novel method, namely integrated knowledge distillation. We propose to align the output of the teacher and student network at a pixel level, which is named output distillation. We further apply a feature-similarity distillation, which intends to transfer the prior information from the feature maps. The similarity matrix is used to represent the rich semantic information from the feature maps. By minimizing the distance of similarity matrices between the teacher and student networks, the distribution of the feature maps of the student network can approach that of the teacher network.

Our contribution can be summarized as follows:

1) We propose an efficient High-Resolution Pose Estimation Network (HRPose) for 6D object pose estimation, which can achieve comparable performance but has about 33\% parameters compared with the state-of-the-art methods on the widely-used LINEMOD dataset.

2) To further improve the accuracy of HRPose, we propose an integrated knowledge distillation method, which transfers both the structure information from the outputs and the feature maps of a trained teacher network, for achieving a mean gain of 1.66\% on accuracy in the average distance of model points metric.

3) Our approach is highly accurate and fast enough (33ms per image) to achieve the speed requirement in real-time tasks.

\section{The Related Work}
In this section, we review related works on RGB-based 6D object pose estimation and knowledge distillation.

\subsection{6D object pose estimation}
Recently, the estimation of 6D object poses including 3D locations and 3D orientations has been an active topic. Previous methods mainly rely on matching techniques\upcite{hinterstoisser2011gradient} or local feature descriptors\upcite{lowe1999object,rublee2011orb} which are not robust to variations of appearances and environments .

Similar to other computer vision tasks, learning-based methods have achieved significant progresses. Given an image, some previous works\upcite{Yu2018PoseCNN,kehl2017ssd} rely on the power of deep neural networks and directly estimate object poses in a single shot. However, the direct regression of 6D poses is still difficult due to the non-linearity of the rotation space\upcite{Peng2020pvnet}, which requires a pose-refinement algorithm to get an accurate 6D pose.

Some recent methods\upcite{tekin2018real,zakharov2019dpod,Peng2020pvnet,Song_2020_CVPR} first predict 2D keypoints of objects and then compute 6D poses through 2D-3D correspondences with a PnP algorithm. In other words, the problem of 6D pose estimation is transformed into the problem of keypoint detections. In this kind of methods, BB8\upcite{rad2017bb8} detects the objects of interest using segmentation and then predicts 2D keypoint coordinates from detected regions. PVNet\upcite{Peng2020pvnet} is proposed to use pixel-wise unit vectors as a representation of keypoints and use the predicted vectors to vote for keypoint locations through RANSAC\upcite{fischler1981random}. DPOD\upcite{zakharov2019dpod} estimates dense multi-class 2D-3D correspondence maps between an input image and available 3D models. HybirdPose\upcite{Song_2020_CVPR} utilizes keypoints, edge vectors, and symmetry correspondences as the representation of 6D poses. HybirdPose achieves a state-of-the-art performance with an additional refinement sub-module.

Despite the accuracy of CNN on pose estimation increases, this relies on its large scalability and time-consuming computations (e.g., VGG\upcite{simonyan2014very} or ResNet\upcite{he2016deep}) which ignores model efficiencies. A few recent works focus on improving the efficiency. Tekin \textit{et al.}\upcite{tekin2018real} employed a lightweight detector YOLOv2\upcite{redmon2017yolo9000} to this task and achieved almost the fastest speed of estimation. However, this method made predictions based on a low-resolution feature map and was not sufficiently precise to meet the accuracy requirement in actual scenarios.

\subsection{Knowledge distillation}
CNNs are expensive in terms of computations and memories. Deeper networks are preferred for accuracy, while smaller networks are widely used due to their efficiency\upcite{zhang2019monocular,jia2020deep,guan2021feature}. So model compression becomes a focus, which intends to speed up running times while maintaining accuracies. Knowledge distillation\upcite{hinton2015distilling} is one of the model compression methods, which transfers knowledge from an accurate teacher network to a compact student network. By utilizing extra supervision information of a trained teacher network, the student network can achieve a better performance.

Bucila \textit{et al.}\upcite{ba2014deep} proposed an algorithm to train a single small neural network by mimicking the output of an ensemble of models. Hinton\upcite{hinton2015distilling} proposed a knowledge distillation (KD) method using softmax outputs of a teacher network as an extra supervision. Since the dimension of both outputs is identical, such an output distillation method can be applied to any pair of networks.

For better utilizing the information contained in the teacher network, some feature distillation methods\upcite{romero2014fitnets, zagoruyko2016paying, yim2017gift,liu2020structured} have been proposed, which transfer intermediate feature representations. Romero \textit{et al.}\upcite{romero2014fitnets} proposed a hint learning method that aligns the intermediate feature maps between the teacher and student network. Zagoruyko \textit{et al.}\upcite{zagoruyko2016paying} proposed to force the student network to mimic the attention maps of a powerful teacher network. Liu \textit{et al.}\upcite{liu2020structured} proposed to distill the pixel-level and structure information from the teacher network simultaneously. Such feature distillation schemes can be combined with an output distillation to improve the performance of the student network.

\section{High-Resolution 6D Pose Estimation Network Using Knowledge Distillation}
Given an RGB image, the task of 6D pose estimation is to detect objects and estimate their 6D poses. The 6D pose can be denoted as a rigid transformation $[\bm{R},\bm{t};0,1]$ from the object coordinate to the camera coordinate, where $\bm{R}$ is a $3 \times 3$ rotation matrix and $\bm{t}$ is a $3 \times 1$ translation vector.

\begin{figure}[!htbp]
	\centering
	\includegraphics[width=0.5\textwidth]{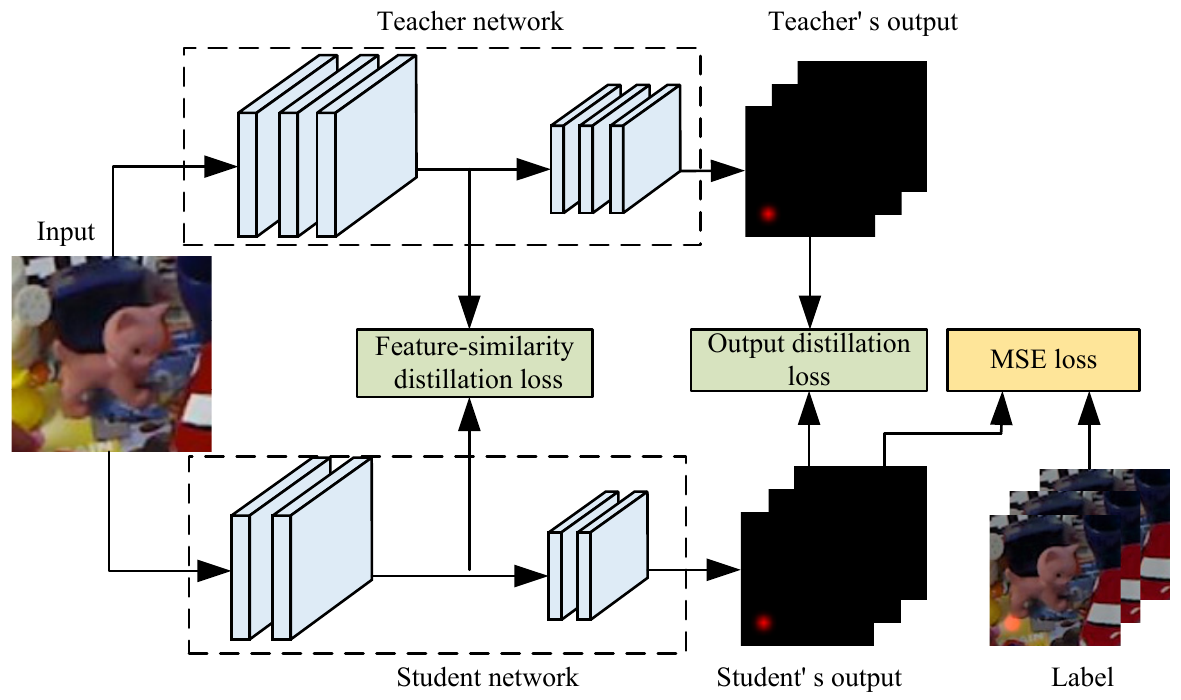}
	\caption{An overview of the proposed HRPose with knowledge distillation. In the training process, we keep the large teacher network fixed and only optimize the student network. The student network is trained with two distillation terms (feature-similarity loss and output distillation loss) and the pose estimation loss. The trained student network can perform an efficient object pose estimation}
	\label{fig:kdstructure}
\end{figure}

We propose a framework of HRPose with knowledge distillation for real-time 6D object pose estimation as shown in Fig.\ref{fig:kdstructure}. We first train a large teacher network that shares the same architecture as the proposed HRPose. Then, we train the proposed HRPose with the assistance of knowledge learned from the teacher network. HRPose takes the small HRNetV2-W18\upcite{wang2020deep} as the backbone, which has fewer convolutional layers than the teacher network. Knowledge distillation happens in this step which transfers both the knowledge of the output and the feature maps from the teacher network to the HRPose.

In this section, we start with an introduction to the proposed HRPose and then describe the details of the knowledge distillation.

\subsection{High-resolution pose estimation network}\par
\begin{figure*}[htbp]
	\centering
	\includegraphics[width=1.0\textwidth]{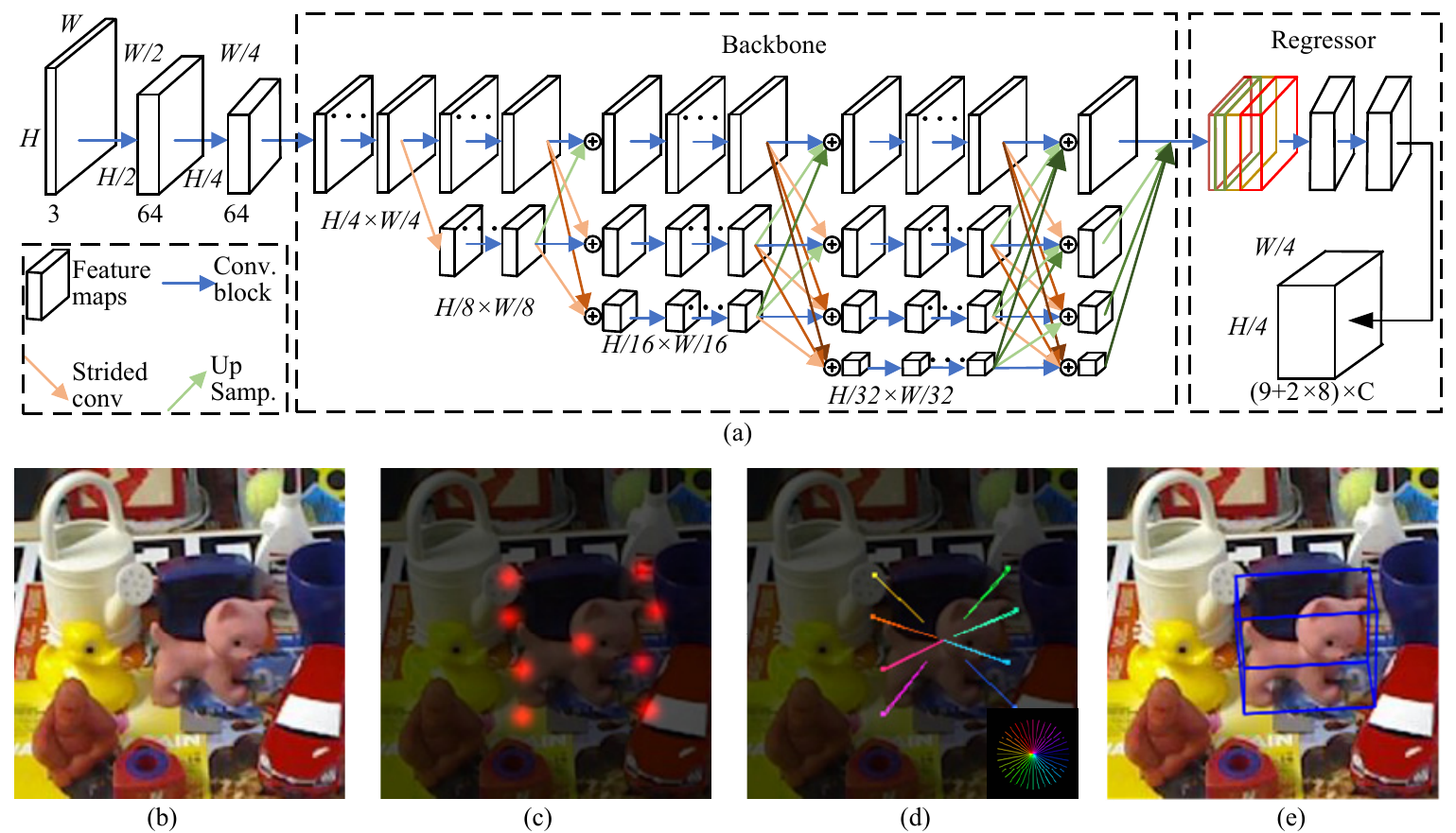}
	\caption{Overview: (a) The architecture of HRPose. In the output of the network, the first nine heatmaps represent the predicted belief maps for keypoints and the latter 16 heatmaps represent the predicted vector fields; (b) An image in the LINEMOD dataset; (c) Belief maps; (d) Vector fields; (e) Predictions of the 2D location of the corners of the projected 3D bounding boxes in the image }\label{fig:networkarch}
\end{figure*}

We propose a two-step pipeline for object pose estimation: we first detect 2D object keypoints using CNNs in a bottom-up method as shown in Fig.\ref{fig:networkarch} and then calculate 6D poses via a PnP algorithm. We select the 8 vertices of the 3D bounding box and the centroid of the object as keypoints.

Given an RGB image size of $H\times W\times 3$, HRPose processes it using a fully-convolutional architecture and predicts a set of 2D belief maps $\bm{H}$ of keypoint locations (Fig.\ref{fig:networkarch}(c)) and a set of 2D vector fields $\bm{L}$, which represents the degree of correlation between the corners and the centriod (Fig.\ref{fig:networkarch}(d)) for each object. The set ${\bm{H}=(\bm{H}_1,\bm{H}_2,...,\bm{H}_I)}$ has $I$ belief maps with a size ${\frac{H}{4}}\times{\frac{W}{4}}$, where each belief map $\bm{H}_i$ represents the location confidence of the $i$-th keypoint. The set ${\bm{L}=(\bm{L}_1,\bm{L}_2,...,\bm{L}_J)}$ has $J$ vector fields with a size ${\frac{H}{4}}\times{\frac{W}{4}\times 2}$. Each pair of vertex and the centroid generate a vector field, where each image location in $\bm{L}_{j}$ denotes a 2D vector. Finally, the belief maps and the vector fields are parsed using a post-processing algorithm to output the locations of 2D keypoints.

The network of HRPose, shown in Fig.\ref{fig:networkarch}(a), consists of three parts: a stem composed of two $3\times 3$ strided convolutions decreasing the resolution to $\frac{1}{4}$, a backbone network that extracts semantic features, and a regressor using the feature of the backbone to estimate the belief maps and the vector fields. The regressor consists of three $1\times 1$ convolutions, and outputs a $\frac{H}{4}\times \frac{W}{4} \times (K\times C)$ tensor representing belief maps and a $\frac{H}{4}\times\frac{W}{4}\times((K-1)\times 2\times C)$ tensor representing vector fields. Here, $C$ and $K$ denote the number of the classes of objects and keypoints for each object, respectively. Considering the positive impact of a high-resolution representation on the keypoint detection, we adopt the small HRNetV2-W18\upcite{wang2020deep} as the backbone, which can maintain a high-resolution representation to achieve a precise keypoint location.

An individual belief maps $\bm{H}_{i}$ is generated by centering a Gaussian kernel around the labeled position. The value at a location $\bm{p}\in \mathbb{R}^2$ in $\bm{H}_{i}$ is defined as,
\begin{equation}
	\bm{H}_{i}\left(\bm{p}\right)=\exp \left(\frac{-{\left\|\bm{p}-\bm{x}_i\right\|}_2^2}{2\sigma^2}\right)
	\label{beliefmaploss}
\end{equation}
where $\bm{x}_i \in \mathbb{R}^2$ denotes the ground-truth position of the $i$-th keypoint in the image and $\sigma$ denotes the standard deviation.

We use a unit vector $\bm{v}_{j}(\bm{p})$ to represent the direction from the $j$-th vertex to the centroid of the corresponding object. Let $\bm{x}_{j}$, $\bm{x}_{\rm{cen}}$ denote the ground-truth position of the $j$-th vertex of the 3D bounding box and the centroid in the image, respectively. If a point $\bm{p}$ lies around the $j$-th vertex, the value at $\bm{L}_j(\bm{p})$ is a unit vector that points from the vertex to the centroid; for all other points, the vector is zero-valued. Therefore, the value at the location $\bm{p}$ in the ground-truth vector field $\bm{L}_j$ is defined as,
\begin{equation}\label{vectorfieldloss}
	\bm{L}_j(\bm{p})=
	\begin{cases}
		\bm{v}_j(\bm{p}),\ \text{if $\bm{p} \in \mathcal{N}(\bm{x}_j)$ or $\bm{p}=\bm{x}_j$}\\
		\bm{0},\ \text{otherwise}
	\end{cases}
\end{equation}
Here, $\bm{v}_j(\bm{p})={\left(\bm{x}_j-\bm{x}_{\rm{cen}}\right)}/{\left\|\bm{x}_j-\bm{x}_{\rm{cen}}\right\|_{2}}$ is the unit vector. $\mathcal{N}(\bm{x}_j)$ denotes the local neighborhood containing pixels within a 3-pixel radius of the ground-truth vertex $\bm{x}_j$.

We use the mean squared error for learning the belief maps and the vector fields. The overall objective loss function is defined as,
\begin{equation}\label{MSEloss} \mathcal{L}_{\rm{mse}}=\frac{1}{I}\sum_{i=1}^{I}\left\|\bm{H}_{i}-\bm{\tilde{H}}_{i}\right\|_{2}^{2}+\frac{1}{J}\sum_{j=1}^{J}\left\|\bm{L}_{j}-\bm{\tilde{L}}_{j}\right\|_{2}^{2}
\end{equation}
where $\bm{H}_{i}$ and $\bm{\tilde{H}}_{i}$ are the $i$-th ground-truth and the predicted belief map, respectively. $\bm{L}_{j}$ and $\bm{\tilde{L}}_{j}$ are the $j$-th ground-truth and predicted vector fields, respectively.

After processing an input image with the proposed network, we can extract 2D keypoint positions from the estimated belief maps using a greedy inference algorithm\upcite{tremblay2018deep}. Since each belief map represents the keypoints of an unknown number of instances of the same type, it is necessary to assemble the detected keypoints to form individual objects according to the predicted vector fields. We take the local peaks from the predicted belief maps above a threshold as keypoints and then group the keypoints into object instances according to the predicted vector fields. For each vertex, we then compare the predicted vector field with the direction from the vertex to the object centroid and assign the detected vertex to the closest object centroid within a certain angular threshold.

When the vertices of each object instance are detected, a PnP\upcite{lepetit2009epnp} algorithm can use the camera intrinsic, the 3D keypoints, and the corresponding projected vertices to compute the final 6D pose.

\subsection{Integrated knowledge distillation}\par
To further improve the pose estimation accuracy without a model performance degradation, we introduce and integrate the knowledge distillation technique, named Integrated Knowledge Distillation. A brief outline of the training method is shown in Fig.\ref{fig:kdstructure}. We want the student network to learn not only the information provided by the ground-truth labels, but also the finer structure knowledge encoded by the teacher network. Let $\rm{T}$ and $\rm{S}$ denote a teacher network and a student network, respectively.

We adopt the output distillation and the feature-similarity distillation to help the training of HRPose jointly. The purpose of the output distillation is intuitive: if the output of a student is similar to that of the teacher, the performance of the student should be similar to the teacher. Transferring the knowledge from the output layer forces the student network to produce a similar output as that of the teacher which is useful to improve the performance of the student network. Here, mean squared error (MSE) is used as a loss function to measure the divergence between the teacher and student outputs. Therefore, the output distillation loss function is formulated as:
\begin{equation}
	\mathcal{L}_{\rm{od}}=\frac{1}{I}\sum_{i=1}^{I}\left\|\bm{H}_{i}^{s}-\bm{H}_{i}^{t}\right\|_{2}^{2}+\\
	\frac{1}{J}\sum_{j=1}^{J}\left\|\bm{L}_{j}^{s}-\bm{L}_{j}^{t}\right\|_{2}^{2}
	\label{outputdistillationloss}
\end{equation}
Here, $\bm{H}_{i}^{s}$ and $\bm{H}_{i}^{t}$ denote the belief maps for the $i$-th keypoint predicted by the pre-trained teacher model and the in-training student model, respectively. Similarly, $\bm{L}_{j}^{s}$ and $\bm{L}_{j}^{t}$ denote the vector fields for the $j$-th keypoint predicted by the teacher and the student models, respectively.

The feature-similarity distillation aims to transfer more structured information from the teacher network to the student network. Generally, features in certain regions share the same properties related to the task. The trained teacher network has extracted certain features related to the object pose estimation task from the original input. The information from the feature maps of the teacher network is valuable for the student network since it provides the student network with guidance on the keypoint detection. Therefore, we apply the feature-similarity distillation to make the student feature maps similar to that of the teacher.

We use $\bm{F}\in \mathbb{R}^{C\times H\times W}$ to denote the output feature map of a layer in the CNN where $C$ is the total number of channels and $H\times W$ is the spatial dimensions. Let $\bm{F}^t$ and $\bm{F}^s$ denote the feature maps from certain layers of the teacher and student networks, respectively. In our method, we assume that the spatial dimensions of $\bm{F}^t$ and $\bm{F}^s$ are identical. The function loss in the feature-similarity distillation is written as
\begin{equation}\label{featuresimilarityloss}
	\mathcal{L}_{fs}=\left\|\bm{G}_{s}-\bm{G}_{t}\right\|_{F}^{2}
\end{equation}
Here, $\bm{G}\in \mathbb{R}^{HW\times HW}$ is the similarity matrix, with each entry $g_{ij}$ defined as,
\begin{equation}
	g_{ij}=\frac{\bm{f}_{i}^{\rm{T}}\bm{f}_{j}}{\left\|\bm{f}_{i}\right\|^2 \left\|\bm{f}_{j}\right\|^2}
\end{equation}
where $\bm{f}_{i} \in \mathbb{R}^{C}$ denotes a feature vector extracted from the $i$-th$(i=1,2,...,H\times W)$ spatial location of the feature map $\bm{F}$. Each item $g_{ij}$ represents the similarity between the $i$-th feature vector and the $j$-th feature vector. With the help of feature-similarity distillation, the student is trained to minimize the divergence between the student and teacher feature maps. Feature-similarity distillation provides more supervision information for student models. In our experiments, we choose to align the feature maps extracted from the backbone because the abstract semantics makes more sense for keypoint detection.

Therefore, the student network is trained to optimize the following loss function:
\begin{equation}\label{KDloss}
	\mathcal{L}=\mathcal{L}_{\rm{mse}}+\lambda_{1}\mathcal{L}_{\rm{od}}+\lambda_{2}\mathcal{L}_{\rm{fs}}
\end{equation}
Here, $\lambda_{1}$ and $\lambda_{2}$ are tunable parameters to balance the standard MSE loss and the distillation loss.

Fig.\ref{fig:kdstructure} summarizes the training of the knowledge transfer framework. The backbone of the teacher network and the student network are HRNetV2-18 and small HRNetV2-18, respectively. We first train a teacher network to optimize Eq. (\ref{MSEloss}) without any extra loss. Then, we train a target student network to minimize the Eq.(\ref{KDloss}), with the knowledge distillation from the teacher network to the target network being conducted throughout the entire training process. At a test time, we only use the efficient and cost-effective HRPose while throwing away the large teacher network, since the target network already extracts the teacher's knowledge.

\section{Experimental Results and Analysis}
\subsection{Dataset and training strategy}\par
\begin{figure}
	\centering
	\includegraphics[width=0.5\textwidth]{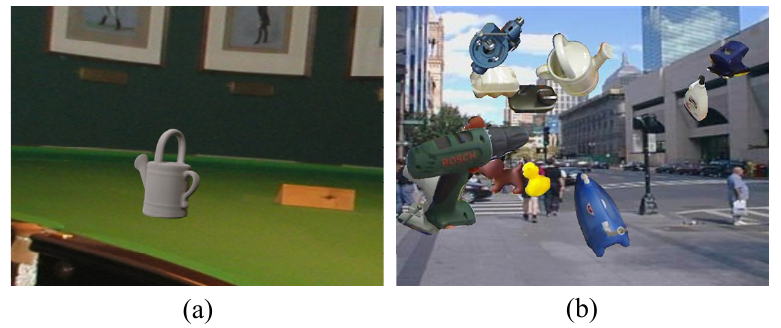}
	\caption{ An illustration of the synthetic images. (a) The rendered image whose pose is uniformly sampled; (b) The synthetic image using ``Cut and Paste''}
	\label{fig:data}
\end{figure}

To validate the proposed method, we perform experiments on the LINEMOD dataset\upcite{hinterstoisser2012model}, which is a standard benchmark for 6D object pose estimation. It provides about 15000 actual images with annotated 6D poses of 13 texture-less objects in heavily cluttered scenes. The precise 3D models of the corresponding objects are also available. We follow prior works\upcite{tekin2018real} to use around 15\% of the LINEMOD examples for training and 85\% for testing. To prevent overfitting, we add synthetic images to the training set following \upcite{Peng2020pvnet}. We render 10000 images for each object and synthesize another 10000 images by the ``Cut and Paste" strategy as shown in Fig.\ref{fig:data}. The background of all synthetic images is randomly sampled from the SUN397 dataset\upcite{SUNdataset}. Besides, we perform online data augmentation including random blur, color jittering, and rotation($\pm 30$ degrees) during training.

In training, we adopt the ADAM optimizer\upcite{kingma2014adam} with a mini-batch size of 32. The initial learning rate is set to 0.0001 and halves every 20 epochs. All models are trained for 120 epochs. Our implementation is based on PyTorch with TITAN XP GPU. In our experiment, We set $\lambda_1$ and $\lambda_2$ to be 0.5 and 0.00005. For simplicity, the proposed teacher network is named ``Teacher", and the student networks with and without knowledge distillation are named ``HRPose+KD" and ``HRPose", respectively.

\subsection{Evaluation metrics}\par
We use two common metrics for evaluation: the Average Distance of Model Points (ADD) metric\upcite{hinterstoisser2012model} and 2D Projection metric\upcite{brachmann2016uncertainty}. The ADD metric is defined as an average distance between the transformed 3D model points using the ground-truth pose and the estimated pose. For the ADD metric, we identify a pose to be correct if the average distance is less than 10\% of the object's diameter. The 2D Projection metric computes the mean distance between the 2D projections of the object's 3D mesh vertices using the estimated and the ground truth pose. A pose is identified to be correct if the distance is less than 5 pixels when using the 2D Projection metric.

Given the ground-truth rotation $\bm{R}$ and the translation $\bm{t}$, the predicted rotation $\tilde{\bm{R}}$ and the translation $\tilde{\bm{t}}$, the ADD metric is calculated as,
\begin{equation}
	\mathrm{ADD}=\frac{1}{m} \sum_{\bm{x} \in \mathcal{M}}\|(\bm{Rx}+\bm{t})-(\tilde{\bm{R}} \bm{x}+\tilde{\bm{t}})\|
\end{equation}
where $\mathcal{M}$ represents the set of 3D model points and $m$ is the number of points. For symmetric objects, the average closest point distance(ADD-S)\upcite{Yu2018PoseCNN} is used to evaluate the performance of 6D pose estimation. The accuracy of pose estimation is defined as the percentage of correct pose estimations. Besides, the number of model parameters and the FLOPs (Floating point Operations) are adopted to evaluate the model efficiency.

\subsection{Comparison with state-of-the-art methods}
We compare the proposed method with the state-of-the-art RGB only methods without any refinement using both ADD metric (shown in Table \ref{tab:LINEMODADD}\upcite{tekin2018real,zakharov2019dpod,Peng2020pvnet,Song_2020_CVPR,Li_2019_ICCV,GDR-Net2021}) and 2D projection error (shown in Table \ref{tab:LINEMOD2d}\upcite{tekin2018real, Peng2020pvnet,GDR-Net2021}). Since some methods do not report their 2D projection accuracy, we do not include them in Table \ref{tab:LINEMOD2d}.

\begin{table*}[!htbp]
	\caption{Quantitative evaluation of 6D pose using ADD(-S) metric on the LINEMOD dataset. The boldface numbers denote the best overall methods. Objects with $*$ are symmetric}\label{tab:LINEMODADD}
	\centering
	\setlength{\tabcolsep}{1.5mm}{
	\begin{tabular}{|c|c|c|c|c|c|c|c|c|c|}
			\hline
			& Tekin\upcite{tekin2018real} & DPOD\upcite{zakharov2019dpod} & PVNet\upcite{Peng2020pvnet} & CDPN\upcite{Li_2019_ICCV} & HybridPose\upcite{Song_2020_CVPR} & GDR-Net\upcite{GDR-Net2021} & Teacher & HRPose & HRPose+KD  \\ \hline
			Ape   & 21.62 & 53.28 & 43.62 & 64.38 & 63.10 & \textbf{76.29} & 68.26 & 61.21 & 65.36(+4.15) \\ \hline
			Benchvise & 81.80 & 95.34 & \textbf{99.90} & 97.77 & \textbf{99.90} & 97.96 & 99.42 & 95.53 & 97.38(+1.85) \\  \hline
			Cam   & 36.57 & 90.36 & 86.86 & 91.67 & 90.40 & \textbf{95.29} & 89.78 & 84.89 & 85.98(+1.09) \\  \hline
			Can   & 68.80 & 94.10 & 95.47 & 95.87 & 98.50 & 98.03 & \textbf{98.62} & 93.60 & 94.88(+1.28) \\  \hline
			Cat   & 41.82 & 60.38 & 79.34 & 83.83 & 89.40 & \textbf{93.21} & 90.02 & 86.03 & 87.33(+1.30) \\ \hline
			Driller & 63.51 & 97.72 & 96.43 & 96.23 & 98.50 & 97.72 & \textbf{98.91} & 96.23 & 96.73(+0.50) \\ \hline
			Duck  & 27.23 & 66.01 & 52.58 & 66.76 & 65.00 & \textbf{80.28} & 72.72 & 67.95 & 71.52(+3..57) \\ \hline
			Eggbox$^{*}$ & 69.58 & 99.72 & 99.15 & 99.72 & \textbf{100.00} & 99.53 & \textbf{100.00} & 98.97 & 99.06(+0.09) \\ \hline
			Glue$^{*}$  & 80.02 & 93.83 & 95.66 & \textbf{99.61} & 98.80 & 98.94 & 98.65 & 97.00 & 97.49(+0.49) \\ \hline
			Holepuncher & 42.63 & 65.83 & 81.92 & 85.82 & 89.70 & \textbf{91.15} & 84.64 & 78.10 & 80.55(+2.45) \\ \hline
			Iron  & 74.97 & 99.80 & 98.88 & 97.85 & \textbf{100.00} & 98.06 & 98.98 & 95.50 & 95.90(+0.40) \\ \hline
			Lamp  & 71.11 & 88.11 & 99.33 & 97.89 & \textbf{99.50} & 99.14 & 99.42 & 96.64 & 97.70(+1.06) \\ \hline
			Phone & 47.74 & 74.24 & 92.41 & 90.75 & \textbf{94.90} & 92.35 & 91.93 & 86.65 & 89.91(+3.26) \\ \hline
			Average & 55.95 & 82.98 & 86.27 & 89.86 & 91.36 & \textbf{93.69} & 91.64 & 87.55 & 89.21(+1.66) \\ \hline
		\end{tabular}%
	}
\end{table*}

\begin{table}[!htbp]	
	\caption{Quantitative evaluation of 6D pose using 2D projection metric on the LINEMOD dataset}\label{tab:LINEMOD2d}
	\centering
	\setlength{\tabcolsep}{0.2mm}{
	\begin{tabular}{|c|c|c|c|c|c|c|}\hline
		&\tabincell{c}{YOLO6D\\ \upcite{tekin2018real}} & \tabincell{c}{PVNet \\ \upcite{Peng2020pvnet}} & \tabincell{c}{GDR-Net\\ \upcite{GDR-Net2021}} & Teacher & HRPose & HRPose+KD \\ \hline
			Ape   & 92.10 & \textbf{99.23} & 98.29 & 98.86 & 97.99 & 98.47(+0.48) \\ \hline
			Benchvise & 95.06 & \textbf{99.81} & 99.32 & 99.03 & 98.35 & 99.13(+0.78) \\ \hline
			Cam   & 93.24 & 99.21 & \textbf{99.41} & \textbf{99.41} & 99.31 & 99.51(+0.20) \\ \hline
			Can   & 97.44 & \textbf{99.90} & 99.51 & 99.70 & 98.3
			3 & 99.02(+0.69) \\ \hline
			Cat   & 97.41 & 99.30 & \textbf{99.60} & 99.30 & 99.20 & 99.30(+0.10) \\ \hline
			Driller & 79.41 & 96.92 & 98.22 & \textbf{98.41} & 97.32 & 98.32(+1.00) \\ \hline
			Duck  & 94.65 & 98.02 & \textbf{98.97} & 98.21 & 98.21 & 98.68(+0.47) \\ \hline
			Eggbox$^*$ & 90.33 & 99.34 & 98.87 & \textbf{99.53} & 99.15 & 99.15(+0.00) \\ \hline
			Glue$^*$  & 96.53 & 98.45 & \textbf{99.42} & 99.23 & 99.23 & 99.42(+0.19) \\ \hline
			Holepuncher & 92.86 & \textbf{100.00} & 99.62 & 99.81 & 97.14 & 97.43(+0.29) \\ \hline
			Iron  & 82.94 & 99.18 & 97.62 & \textbf{99.28} & 96.93 & 97.44(+0.51) \\ \hline
			Lamp  & 76.87 & 98.27 & 96.64 & \textbf{98.66} & 96.64 & 97.79(+1.15) \\ \hline
			Phone & 86.07 & \textbf{99.42} & 97.92 & 99.33 & 97.89 & 98.85(+0.96) \\ \hline
			Average & 90.38 & 99.00 & 98.72 & \textbf{99.14} & 98.13 & 98.67(+0.46) \\ \hline
	\end{tabular}}
\end{table}

Table \ref{tab:LINEMODADD} indicates that the proposed simple ``HRPose" achieves 87.55\% pose estimation accuracy without any extra information on average and outperforms PVNet\upcite{Peng2020pvnet}. With the help of knowledge distillation, the overall pose estimation accuracy of ``HRPose" raises from 87.55\% to 89.21\% in the ADD metric and from 98.13\% to 98.67\% in the 2D projection metric. Although ``HRPose+KD" does not outperform HybridPose\upcite{Song_2020_CVPR}, it still achieves comparable pose estimation accuracy with merely 33\% parameters and 20.6\% FLOPs of HybridPose\upcite{Song_2020_CVPR}, as shown in Table \ref{tab:performance}.

From both Table \ref{tab:LINEMODADD} and Table \ref{tab:LINEMOD2d}, we can observe that the distilled small network achieves a better 6D pose estimation performance than its corresponding baselines using both ADD metric and 2D projection metric. Especially, the ``HRPose+KD" outperforms the baseline model by a significant margin of 4.15\% on ``Ape" using the ADD metric.

\begin{table}
\centering
	\caption{Comparison of different methods in the backbone, model size (the number of model parameters), and computational cost (FLOPs). M/G:$10^6/10^9$}\label{tab:performance}
	\begin{tabular}{|c|c|c|c|}
			\hline
			Methods & Backbone & \#Params & FLOPs \\ \hline
			YOLO6D\upcite{tekin2018real} & YOLOv2 & 50.5M & 26.1G \\ \hline	
			PVNet\upcite{Peng2020pvnet} & ResNet-18 & 12.9M  & 72.7G \\ \hline
			HybridPose\upcite{Song_2020_CVPR} & ResNet-18 & 12.9M & 75.2G \\ \hline
			GDR-Net\upcite{GDR-Net2021} & ResNet-34 & 33.5M & - \\    \hline
			Teacher  & HRNetV2-W18 & 9.7M  & 23.2G \\ \hline
			HRPose & small HRNetV2-W18 & 4.2M  & 15.5G \\ \hline
	\end{tabular}		
\end{table}

\begin{figure}
	\includegraphics[width=0.5\textwidth]{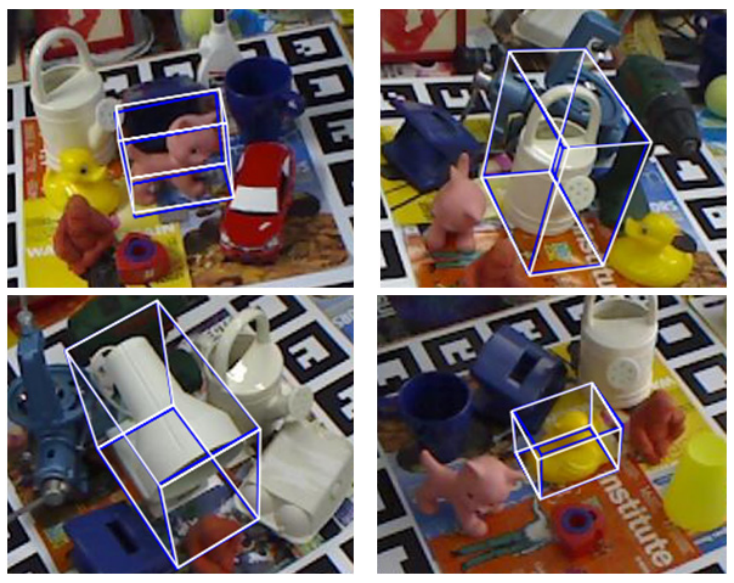}
	\caption{Visualization of results on the LINEMOD dataset. White and blue bounding boxes represent the ground-truth and estimated poses respectively}\label{fig:result}
\end{figure}

Fig.\ref{fig:result} provides some qualitative results on the LINEMOD dataset. It can be observed that HRPose can achieve robust and reliable pose estimation with various background clutters.

We calculate the number of network parameters (\#Params) and the sum of float point operations (FLOPs) to measure the model efficiency. The resolution of the input RGB image is $640\times 480$. As shown in Table \ref{tab:performance}, HRPose has the minimal model size and the lowest computational complexity. Note that, GDR-Net\upcite{GDR-Net2021} needs a detector to obtain the object region and we do not count the model size of the detector. The distilled HRPose model achieves comparable even better results with only about 8.3\% of the model size of YOLO6D, 33\% of the model size of HybridPose\upcite{Song_2020_CVPR}. Although the accuracy of our model is slightly lower than GDR-Net, it is still a comparable result (89.21\% in the ADD metric with 98.67\% in the 2d Projection metric) with almost 13\% of the model size of GDR-Net. This means our model achieves relative comparable results with a cheaper deployment cost. Also, the proposed method can run at 33fps on a GTX 2080 GPU which can satisfy the requirement for real-time object pose estimation.

\subsection{Ablation study}\par
To investigate the effectiveness of different components of our distillation scheme, we conduct an ablation study on the object ``Cat'' from LINEMOD dataset. From Table \ref{tab:ablationstudy}, we can observe that: (i)With the output-distillation ($\mathcal{L}_{\rm{od}}$) and the feature-similarity distillation ($\mathcal{L}_{\rm{fs}}$), it achieves 1.02\% and 0.79\% improvements in term of ADD metric, respectivelly. (ii)With the combination of the output-distillation and the feature-similarity, the proposed model achieves an improvement of 1.30\%(86.03\%-87.33\%) accuracy. These observations indicate that the two distillation schemes that we present can improve the accuracy of the network, which can be combined to help the student network to obtain a better performance.

\begin{table}
\centering
\caption{Ablation study of different components of the loss in the proposed method. $\mathcal{L}_{\rm{od}}$: output distillation; $\mathcal{L}_{\rm{fs}}$: feature-similarity distillation}\label{tab:ablationstudy}
\setlength{\tabcolsep}{0.5mm}{
	\begin{tabular}{|c|c|c|c|c|}
		\hline
		& W/O distillation & $\mathcal{L}_{\rm{od}}$ & $\mathcal{L}_{\rm{fs}}$ & $\mathcal{L}_{\rm{od}}$+$\mathcal{L}_{\rm{fs}}$ \\ \hline
		ADD & 86.03 & 87.05(+1.02) & 86.83(+0.80)  & 87.33(+1.30) \\ \hline
	\end{tabular}}
\end{table}

Besides, we perform an ablation study on the setting of the hyperparameters $\lambda_1$,$\lambda_2$. For simplicity, we fix $\lambda_1$ to be 0.5. Table \ref{tab:paramablationstudy} reports the impact of the hyperparameters on the training process using ADD metric, where $\lambda_2$ increases from 0.00005 to 0.001. Then we fix $\lambda_2$ to be 0.00005 and let $\lambda_1$ vary from 0.05 to 1. It can be observed that the proposed HRPose achieves a higher accuracy varying from 0.00005 to 0.0005 compared with the ADD accuracy obtained with $\lambda_2=0$. When $\lambda_1=0.5$ and $\lambda_2=0.00005$, the proposed model achieves the highest accuracy. However, if the setting of the hyperparameters is too large (e.g. $\lambda_1=1$ or $\lambda_2$=0.001), knowledge distillation will disrupt the training of the student network which leads to the failure of convergence.

\begin{table}
\centering
	\caption{Ablation study on the selection of hyperparameters $\lambda$}\label{tab:paramablationstudy}
	\begin{tabular}{|c|c|c|c|c|c|}
		\hline
		\multirow{2}{*}{$\lambda_1=0.5$} & \multicolumn{5}{|c|}{$\lambda_2$} \\
		\cline{2-6}
		& 0     & 0.00005 & 0.0001 & 0.0005 & 0.001 \\ \hline
		ADD   & 87.05 & \textbf{87.33} & 87.23 & 87.14 & 85.73 \\ \hline
		\multirow{2}{*}{$\lambda_2=0.00005$} & \multicolumn{5}{|c|}{$\lambda_1$} \\
		\cline{2-6}
		& 0     & 0.05  & 0.1   & 0.5   & 1 \\ \hline
		ADD   & 86.83 & 87.23 & 87.14 & \textbf{87.33} & 86.03 \\ \hline
	\end{tabular}
\end{table}

\section{Conclusion}
In this paper, we have proposed a simple and lightweight High-Resolution 6D Pose Estimation Network (HRPose) by adopting the small HRNet as a feature extractor. This design is helpful to reduce computation burdens while guaranteeing a high accuracy of pose estimation. With about 33\% parameters of the state-of-the-art models, our HRPose can achieve comparable performance on the widely-used benchmark LINEMOD dataset. To enhance the performance of HRPose, we have also proposed a novel knowledge distillation technique that transfers the structure knowledge from a large and complex network to the proposed HRPose. With the help of the proposed knowledge distillation method, the performance of the proposed HRPose can be further improved for 6D object pose estimation. Our method is highly accurate and fast enough (33 frames per second) to satisfy the real-time requirement.

\vspace{11pt}
\begin{IEEEbiography}[{\includegraphics[width=1in,height=1.25in,clip,keepaspectratio]{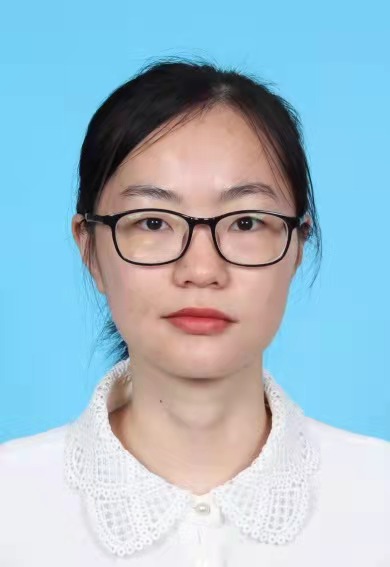}}]{Qi Guan received the B.S. degree in measurement and control from Southeast University, Nanjing, P. R. China, in 2019. She is currently pursuing the M.S. degree in control engineering with Shanghai Jiao Tong University, Shanghai, P. R. China. Her research interests are 6D pose estimation and real-time application in deep learning.}
\end{IEEEbiography}

\begin{IEEEbiography}[{\includegraphics[width=1in,height=1.25in,clip,keepaspectratio]{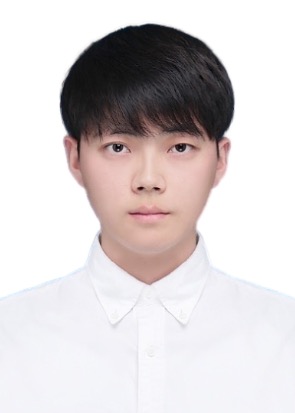}}]{Zihao Sheng received the B.S. degree in automation from Xi¡¯an Jiaotong University, Xi¡¯an, P. R. China, in 2019. He is currently pursuing the M.S. degree in control engineering with Shanghai Jiao Tong University, Shanghai, P. R. China. His research interests include intelligent transportation systems, autonomous driving, and intelligent control. }
\end{IEEEbiography}

\begin{IEEEbiography}[{\includegraphics[width=1in,height=1.25in,clip,keepaspectratio]{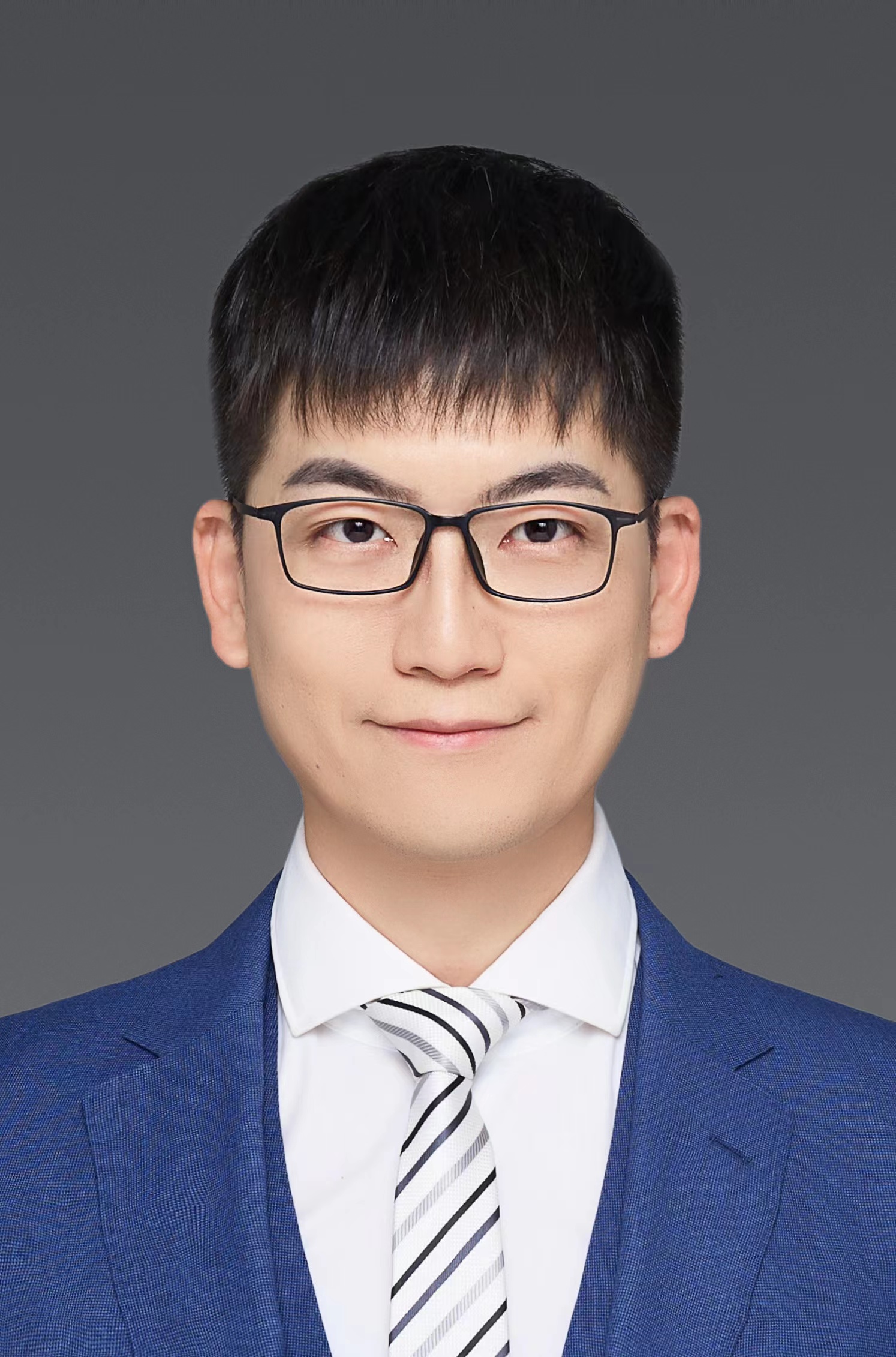}}]{Shibei Xue received the Ph.D. degree in control science and engineering from Tsinghua University, Beijing, P. R. China, in 2013.From 2014 to 2016, he was a Post-Doctoral Researcher with the University of New South Wales, Canberra, ACT, Australia, and then, he worked as a Post-Doctoral Researcher with the Department of Physics, National Cheng Kung University, Taiwan, P. R. China. In July 2017, he joined Shanghai Jiao Tong University, Shanghai, P. R. China, where he is currently an Associate Professor with the Department of Automation. He was selected for the Shanghai Pujiang Program funded by the Shanghai Science and Technology Committee in 2018. His research interests include quantum control, optimization and intelligent control of complex systems.\ \\(Email:shbxue@sjtu.edu.cn)}
\end{IEEEbiography}

\vspace{11pt}
\vfill

\end{document}